\documentclass{article}

\usepackage{spconf}
\usepackage{amsmath,amssymb,amsthm,mathtools,amsfonts}
\usepackage{graphicx}
\usepackage{subcaption}
\usepackage{hyperref}
\usepackage{url}
\usepackage{booktabs}
\usepackage{nicefrac}
\usepackage{microtype}
\usepackage{xcolor}
\usepackage{xspace}
\usepackage{placeins}
\usepackage{float}
\usepackage{array}
\usepackage{tabularx}
\usepackage{multirow}
\usepackage{enumitem}
\usepackage{tikz}
\usepackage{wrapfig}
\usepackage{etoolbox}

\definecolor{turquoise}{HTML}{00A6A6}
\definecolor{resultgreen}{HTML}{008000}
\colorlet{BROWN}{brown}

\newif\ifshowcolors
\showcolorsfalse
\newcommand{\mycolor}[1]{\ifshowcolors\color{#1}\else\color{.}\fi}

\newcommand{\Sref}[2][]{\hyperref[#2]{Sec.~\ref*{#2}#1}}
\newcommand{\Fref}[2][]{\hyperref[#2]{Fig.~\ref*{#2}#1}}
\newcommand{\Tref}[2][]{\hyperref[#2]{Tab.~\ref*{#2}#1}}
\newcommand{\Eref}[2][]{\hyperref[#2]{Eq.~\ref*{#2}#1}}
\newcommand{\Aref}[2][]{\hyperref[#2]{Appx.~\ref*{#2}#1}}
\newcommand{\Dref}[2][]{\hyperref[#2]{Def.~\ref*{#2}#1}}

\newcommand{\plainurl}[1]{\href{#1}{#1}}

\makeatletter
\apptocmd{\thebibliography}{%
    \setlength{\itemsep}{0pt}%
    \setlength{\parsep}{0pt}%
    \setlength{\emergencystretch}{1em}%
    \Urlmuskip=0mu plus 1mu\relax}{}{}
\makeatother

\theoremstyle{definition}

\hypersetup{
    colorlinks,
    linkcolor={red!50!black},
    citecolor={red!50!black},
    urlcolor={red!50!black}
}

\title{{\mycolor{brown} Causal Analysis and Mitigation of\\
Spurious Onsets in Full-Duplex Speech LLMs}}

\name{Kento Nishi$^{\star\dagger}$}
\address{$^{\star}$Massachusetts Institute of Technology \qquad
$^{\dagger}$Comcast Applied AI Research, Speech AI Team}

\begin{document}
\ninept
\setlength{\textfloatsep}{8pt plus 2pt minus 2pt}
\setlength{\abovedisplayskip}{1pt}
\setlength{\belowdisplayskip}{1pt}
\setlength{\abovedisplayshortskip}{1pt}
\setlength{\belowdisplayshortskip}{1pt}
\maketitle

\begin{abstract}
\setlength{\baselineskip}{9.7pt}
Speech-to-speech LLMs like Moshi, and its derivative PersonaPlex, can listen and speak concurrently through full-duplex generation.
However, they can begin speaking inappropriately during prolonged user silence: under digital-zero input, Moshi and PersonaPlex initiate speech in 30\% and 27.5\% of five-minute continuations, respectively.
What causes this spurious speech?
We investigate two hypotheses: either repeated sampling selects speech despite persistently low onset probabilities, or self-conditioning on nonspeech outputs causes an abrupt spike in onset probability.
We find that, at every observed onset, speech probability spikes by over nine orders of magnitude in one 80-ms frame, supporting the latter hypothesis.
Then, to suppress these onsets without blocking genuine responses, we ask a causal counterfactual question: is the model responding to user speech, or would its next-token distribution remain similar if the preceding user input were muted?
Accordingly, we suppress onsets whose distributions change little under this intervention.
Under realistic microphone noise, our method suppresses spurious onsets, while preserving genuine responses: one-sided 95\% lower confidence bounds are 98.68\% and 98.82\% for Moshi, and 96.90\% and 99.25\% for PersonaPlex.
Our inference-time method runs in real-time without retraining, with 95th-percentile decision time below 61 ms, within the 80-ms frame budget.
Our code is available at \mbox{\plainurl{https://github.com/KentoNishi/icassp27-spurious-onsets}}.
\end{abstract}


\section{Introduction}
\label{sec:introduction}

Most spoken dialogue systems have explicit turn boundaries. That is to say, a voice activity detector (VAD) closes the user's ``turn,'' an automatic speech recognition service (ASR) produces a transcript, a large language model (LLM) generates a response, and a text-to-speech synthesizer (TTS) renders the response as audio. Streaming components can overlap some of these computations, but the cascaded architecture still converts spoken interaction into alternating segments of listening and speaking. As a result, contemporary systems cannot respond while the user is speaking, handle interruptions, or understand and produce overlapping speech. Passing through text also discards non-linguistic information that can affect meaning, including prosody, emotion, laughter, and other non-speech vocalizations. Moreover, the sequence of independent components introduces additional latency between the user and the model~\cite{defossez2024moshi,veluri2024syncllm,lin2025fullduplexbench}.

\usetikzlibrary{calc,arrows.meta,positioning}

\definecolor{panelgray}{RGB}{242,242,242}
\definecolor{tilefill}{RGB}{250,250,250}
\definecolor{predblue}{HTML}{1F77B4}
\definecolor{datagray}{HTML}{666666}
\definecolor{modelred}{HTML}{D62728}
\definecolor{nvidiagreen}{HTML}{76B900}
\definecolor{emitred}{HTML}{D62728}
\definecolor{baselinegray}{RGB}{150,150,150}

\tikzset{
  flow/.style={draw=black,line width=1.55pt,line cap=round,line join=round,-{Latex[length=2.8mm,width=2.6mm]},shorten <=1pt,shorten >=1pt},
  flowc/.style={line width=1.55pt,line cap=round,line join=round,rounded corners=5pt,-{Latex[length=2.8mm,width=2.6mm]},shorten <=1pt,shorten >=1pt},
  panel/.style={draw=black,line width=0.8pt,fill=tilefill},
  badge/.style={scale=\BadgeScale,transform shape,rounded corners=6.7pt,inner xsep=2.7pt,inner ysep=4pt,font=\fontsize{8.5}{8.7}\sffamily\bfseries,text=white},
  lab/.style={font=\fontsize{9}{9.2}\sffamily\bfseries,anchor=west},
  plab/.style={font=\fontsize{10}{10}\selectfont},
}

\def\PanelSize{2.6}
\def\HalfPanel{1.3}
\def\LaneTop{0.45}
\def\LaneBot{-0.45}

\newif\iflabelsonlyleftmost
\labelsonlyleftmosttrue

\def\UserIconScale{1}
\def\UserIconX{-1.04}
\def\UserIconY{0.36}
\def\UserLabelX{-0.92}
\def\UserLabelY{0.37}

\def\ModelIconScale{1}
\def\ModelIconX{-1.04}
\def\ModelIconY{0.40}
\def\ModelLabelX{-0.92}
\def\ModelLabelY{0.43}

\def\MuteIconScale{1.30}
\def\SpeakIconScale{1.30}
\def\SpeechBarColor{modelred}

\def\BadgeScale{0.81}
\def\BadgeInsetX{0.08}
\def\BadgeInsetY{0.08}
\def\ObservedBadgeCorner{ne}
\def\MutedBadgeCorner{ne}
\def\SuppressBadgeCorner{se}
\def\EmitBadgeCorner{se}
\def\SuppressBadgeTextYShift{-0.35pt}
\newcommand{\ShiftTextDown}[2]{%
  \vphantom{#1}%
  \smash{\raisebox{#2}{#1}}%
}

\def\StartX{-0.98}
\def\AxisLen{1.96}

\def\GreetingModelOffset{0.05}
\def\GreetingModelLen{0.34}
\def\HistoryUserOffset{0.43}
\def\HistoryUserLen{0.38}
\def\HistoryModelOffset{0.84}
\def\HistoryModelLen{0.44}
\def\NewUserOffset{1.32}
\def\NewUserLen{0.36}
\def\OnsetOffset{1.64}
\def\OnsetLen{0.28}

\def\UserHistoryOpacity{1.00}
\def\ModelHistoryOpacity{0.34}
\def\IncomingOpacity{0.34}
\def\ActiveOpacity{1.00}

\def\ArrowStub{0.55}
\def\CompareHalf{0.98}
\def\EqYOffset{0.40}
\def\MathInset{0.16}

\def\EqTextYShift{-0.7pt}

\def\FrameNoteHalfWidth{0.7}
\def\FrameNoteXOffset{0.00}
\def\FrameNoteYOffset{0.20}
\def\FrameNoteTextGap{0.09}
\def\FrameNoteFontSize{8.5}

\newcommand{\PersonIcon}[3]{%
\begin{scope}[shift={({#1},{#2})},scale=#3]
  \fill[predblue] (0,0.14) circle (0.085);
  \fill[predblue,rounded corners=0.03] (-0.13,-0.08) rectangle (0.13,0.07);
\end{scope}}

\newcommand{\RobotIcon}[3]{%
\begin{scope}[shift={({#1},{#2})},scale=#3]
  \draw[modelred,line width=0.045cm,rounded corners=0.035cm,fill=white] (-0.13,-0.08) rectangle (0.13,0.10);
  \fill[modelred] (-0.05,0.01) circle (0.016);
  \fill[modelred] ( 0.05,0.01) circle (0.016);
  \draw[modelred,line width=0.038cm] (0,0.10)--(0,0.16);
  \fill[modelred] (0,0.18) circle (0.016);
\end{scope}}

\newcommand{\MuteIcon}[3]{%
\begin{scope}[shift={({#1},{#2})},scale=#3]
  \fill[datagray] (-0.10,-0.05) rectangle (-0.03,0.05);
  \fill[datagray] (-0.03,-0.09)--(0.07,-0.14)--(0.07,0.14)--(-0.03,0.09)--cycle;
  \draw[datagray,line width=0.032cm] (-0.11,-0.13)--(0.14,0.12);
\end{scope}}

\newcommand{\SpeakIcon}[3]{%
\begin{scope}[shift={({#1},{#2})},scale=#3]
  \fill[\SpeechBarColor] (-0.10,-0.05) rectangle (-0.03,0.05);
  \fill[\SpeechBarColor] (-0.03,-0.09)--(0.07,-0.14)--(0.07,0.14)--(-0.03,0.09)--cycle;
  \draw[\SpeechBarColor,line width=0.028cm] (0.11,-0.07) arc[start angle=-40,end angle=40,radius=0.10];
  \draw[\SpeechBarColor,line width=0.025cm] (0.16,-0.10) arc[start angle=-40,end angle=40,radius=0.15];
\end{scope}}

\newcommand{\Baseline}[3]{\draw[baselinegray,line width=0.28pt] (#1,#2)--++(#3,0);}

\newcommand{\SilenceMarks}[3]{%
  \foreach \u in {0.14,0.30,0.46,0.62,0.78,0.94}{%
    \draw[baselinegray,line width=0.22pt] ({#1+\u*#3-0.022},{#2-0.022})--({#1+\u*#3+0.022},{#2+0.022});}}

\newcommand{\AudioAxis}[3]{%
  \Baseline{#1}{#2}{#3}
  \SilenceMarks{#1}{#2}{#3}}

\newcommand{\WaveAOpacity}[5]{%
\begin{scope}[shift={({#1},{#2})},opacity=#5]
  \begin{scope}[xscale=#3/1.20,yscale=0.27/0.34]
    \fill[#4] (0.00,0.00)--(0.04,0.03)--(0.08,0.08)--(0.12,0.16)--(0.16,0.10)--(0.20,0.24)--(0.24,0.14)--(0.28,0.30)--(0.32,0.18)--(0.36,0.28)--(0.40,0.15)--(0.44,0.34)--(0.48,0.21)--(0.52,0.29)--(0.56,0.17)--(0.60,0.31)--(0.64,0.19)--(0.68,0.27)--(0.72,0.16)--(0.76,0.25)--(0.80,0.14)--(0.84,0.21)--(0.88,0.12)--(0.92,0.19)--(0.96,0.10)--(1.00,0.16)--(1.04,0.08)--(1.08,0.12)--(1.12,0.05)--(1.16,0.03)--(1.20,0.00)
      --(1.16,-0.03)--(1.12,-0.05)--(1.08,-0.12)--(1.04,-0.08)--(1.00,-0.16)--(0.96,-0.10)--(0.92,-0.19)--(0.88,-0.12)--(0.84,-0.21)--(0.80,-0.14)--(0.76,-0.25)--(0.72,-0.16)--(0.68,-0.27)--(0.64,-0.19)--(0.60,-0.31)--(0.56,-0.17)--(0.52,-0.29)--(0.48,-0.21)--(0.44,-0.34)--(0.40,-0.15)--(0.36,-0.28)--(0.32,-0.18)--(0.28,-0.30)--(0.24,-0.14)--(0.20,-0.24)--(0.16,-0.10)--(0.12,-0.16)--(0.08,-0.08)--(0.04,-0.03)--cycle;
  \end{scope}
\end{scope}}

\newcommand{\WaveBOpacity}[5]{%
\begin{scope}[shift={({#1},{#2})},opacity=#5]
  \begin{scope}[xscale=#3/1.20,yscale=0.27/0.34]
    \fill[#4] (0.00,0.00)--(0.04,0.02)--(0.08,0.06)--(0.12,0.12)--(0.16,0.20)--(0.20,0.14)--(0.24,0.26)--(0.28,0.18)--(0.32,0.32)--(0.36,0.22)--(0.40,0.29)--(0.44,0.17)--(0.48,0.27)--(0.52,0.20)--(0.56,0.34)--(0.60,0.19)--(0.64,0.30)--(0.68,0.16)--(0.72,0.28)--(0.76,0.20)--(0.80,0.25)--(0.84,0.14)--(0.88,0.22)--(0.92,0.11)--(0.96,0.18)--(1.00,0.09)--(1.04,0.14)--(1.08,0.07)--(1.12,0.10)--(1.16,0.04)--(1.20,0.00)
      --(1.16,-0.04)--(1.12,-0.10)--(1.08,-0.07)--(1.04,-0.14)--(1.00,-0.09)--(0.96,-0.18)--(0.92,-0.11)--(0.88,-0.22)--(0.84,-0.14)--(0.80,-0.25)--(0.76,-0.20)--(0.72,-0.28)--(0.68,-0.16)--(0.64,-0.30)--(0.60,-0.19)--(0.56,-0.34)--(0.52,-0.20)--(0.48,-0.27)--(0.44,-0.17)--(0.40,-0.29)--(0.36,-0.22)--(0.32,-0.32)--(0.28,-0.18)--(0.24,-0.26)--(0.20,-0.14)--(0.16,-0.20)--(0.12,-0.12)--(0.08,-0.06)--(0.04,-0.02)--cycle;
  \end{scope}
\end{scope}}

\newcommand{\WaveA}[4]{\WaveAOpacity{#1}{#2}{#3}{#4}{1}}
\newcommand{\WaveB}[4]{\WaveBOpacity{#1}{#2}{#3}{#4}{1}}

\newcommand{\SquarePanel}[2]{\draw[panel] ({#1-\HalfPanel},{#2-\HalfPanel}) rectangle ({#1+\HalfPanel},{#2+\HalfPanel});}

\newcommand{\BadgeNode}[6]{\node[badge,fill=#4,anchor=#5,yshift=#6] at #2 {#3};}
\newcommand{\badgenw}[5]{\BadgeNode{#1}{({#1-\HalfPanel+\BadgeInsetX},{#2+\HalfPanel-\BadgeInsetY})}{#3}{#4}{north west}{#5}}
\newcommand{\badgene}[5]{\BadgeNode{#1}{({#1+\HalfPanel-\BadgeInsetX},{#2+\HalfPanel-\BadgeInsetY})}{#3}{#4}{north east}{#5}}
\newcommand{\badgesw}[5]{\BadgeNode{#1}{({#1-\HalfPanel+\BadgeInsetX},{#2-\HalfPanel+\BadgeInsetY})}{#3}{#4}{south west}{#5}}
\newcommand{\badgese}[5]{\BadgeNode{#1}{({#1+\HalfPanel-\BadgeInsetX},{#2-\HalfPanel+\BadgeInsetY})}{#3}{#4}{south east}{#5}}
\newcommand{\BadgeAt}[6]{%
  \ifcsname badge#5\endcsname
    \csname badge#5\endcsname{#1}{#2}{#3}{#4}{#6}%
  \else
    \PackageError{fig1}{Unknown badge corner '#5'}{Use nw, ne, sw, or se.}%
  \fi}

\newcommand{\UserLabel}[2]{%
  \PersonIcon{#1+\UserIconX}{#2+\UserIconY}{\UserIconScale}
  \node[lab,text=predblue] at ({#1+\UserLabelX},{#2+\UserLabelY}) {user};}

\newcommand{\ModelLabel}[2]{%
  \RobotIcon{#1+\ModelIconX}{#2+\ModelIconY}{\ModelIconScale}
  \node[lab,text=modelred] at ({#1+\ModelLabelX},{#2+\ModelLabelY}) {model};}

\newcommand{\MaybeLabels}[3]{%
  \iflabelsonlyleftmost
    \ifnum#3=1\relax
      \UserLabel{#1}{#2+\LaneTop}
      \ModelLabel{#1}{#2+\LaneBot}
    \fi
  \else
    \UserLabel{#1}{#2+\LaneTop}
    \ModelLabel{#1}{#2+\LaneBot}
  \fi}

\newcommand{\HistoryUserRow}[2]{%
  \AudioAxis{#1+\StartX}{#2+\LaneTop}{\AxisLen}
  \WaveAOpacity
    {#1+\StartX+\HistoryUserOffset}
    {#2+\LaneTop}
    {\HistoryUserLen}
    {predblue}
    {\UserHistoryOpacity}}

\newcommand{\ModelHistoryWaves}[2]{%
  \WaveBOpacity
    {#1+\StartX+\GreetingModelOffset}
    {#2}
    {\GreetingModelLen}
    {modelred}
    {\ModelHistoryOpacity}
  \WaveBOpacity
    {#1+\StartX+\HistoryModelOffset}
    {#2}
    {\HistoryModelLen}
    {modelred}
    {\ModelHistoryOpacity}}

\newcommand{\HistoryModelRow}[2]{%
  \AudioAxis{#1+\StartX}{#2+\LaneBot}{\AxisLen}
  \ModelHistoryWaves{#1}{#2+\LaneBot}}

\newcommand{\NewUserUtteranceOpacity}[3]{%
  \WaveAOpacity
    {#1+\StartX+\NewUserOffset}
    {#2+\LaneTop}
    {\NewUserLen}
    {predblue}
    {#3}}

\newcommand{\NewUserUtterance}[2]{%
  \NewUserUtteranceOpacity{#1}{#2}{\ActiveOpacity}}

\newcommand{\SharedHistoryRows}[3]{%
  \MaybeLabels{#1}{#2}{#3}
  \HistoryUserRow{#1}{#2}
  \HistoryModelRow{#1}{#2}}

\newcommand{\SourcePanel}[2]{%
  \SquarePanel{#1}{#2}
  \SharedHistoryRows{#1}{#2}{1}
  \NewUserUtterance{#1}{#2}}

\newcommand{\ObservedPanel}[4]{%
  \SquarePanel{#1}{#2}
  \BadgeAt{#1}{#2}{#3}{predblue}{#4}{0pt}
  \SharedHistoryRows{#1}{#2}{0}
  \NewUserUtterance{#1}{#2}}

\newcommand{\MutedPanel}[4]{%
  \SquarePanel{#1}{#2}
  \BadgeAt{#1}{#2}{#3}{datagray}{#4}{0pt}
  \SharedHistoryRows{#1}{#2}{0}
  \NewUserUtteranceOpacity{#1}{#2}{\IncomingOpacity}}

\newcommand{\SuppressPanel}[4]{%
  \SquarePanel{#1}{#2}
  \BadgeAt
  {#1}{#2}
  {\ShiftTextDown{#3}{-1pt}}
  {datagray}{#4}{\SuppressBadgeTextYShift}
  \AudioAxis{#1+\StartX}{#2}{\AxisLen}
  \ModelHistoryWaves{#1}{#2}}

\newcommand{\EmitPanel}[4]{%
  \SquarePanel{#1}{#2}
  \BadgeAt{#1}{#2}{#3}{emitred}{#4}{0pt}
  \AudioAxis{#1+\StartX}{#2}{\AxisLen}
  \ModelHistoryWaves{#1}{#2}
  \WaveAOpacity
    {#1+\StartX+\OnsetOffset}
    {#2}
    {\OnsetLen}
    {emitred}
    {\ActiveOpacity}}

\newcommand{\ProbabilityPanel}[5]{%
  \SquarePanel{#1}{#2}
  \def\barBaseL{0.78}
  \def\barCtrL{-0.28}
  \def\barCtrR{0.28}
  \draw[line width=0.42pt] ({#1-\barBaseL},{#2-0.46})--({#1+\barBaseL},{#2-0.46});
  #5
  \MuteIcon{#1+\barCtrL}{#2-0.79}{\MuteIconScale}
  \SpeakIcon{#1+\barCtrR}{#2-0.79}{\SpeakIconScale}
  \node[plab,text=#4,anchor=north east] at ({#1+\HalfPanel-\MathInset},{#2+\HalfPanel-\MathInset}) {$\mathbf{#3}$};}

\newcommand{\FrameLatencyNote}[2]{%
  \pgfmathsetmacro{\FrameNoteX}
    {#1+\FrameNoteXOffset}

  \pgfmathsetmacro{\FrameNoteY}
    {#2-\CompareHalf-\FrameNoteYOffset}

  \draw[
    black!60,
    line width=0.45pt
  ]
    ({\FrameNoteX-\FrameNoteHalfWidth},{\FrameNoteY})
    --
    ({\FrameNoteX+\FrameNoteHalfWidth},{\FrameNoteY});

  \node[
    anchor=north,
    text=black!65,
    font=\fontsize{\FrameNoteFontSize}{\FrameNoteFontSize}\sffamily
  ]
    at
    ({\FrameNoteX},{\FrameNoteY-\FrameNoteTextGap})
    {1 frame = 80 ms};
}

\newcommand{\PBars}[2]{%
  \fill[datagray!75] ({#1-0.28-0.19},{#2-0.46}) rectangle ({#1-0.28+0.19},{#2+0.02});
  \fill[\SpeechBarColor] ({#1+0.28-0.19},{#2-0.46}) rectangle ({#1+0.28+0.19},{#2+0.88});}

\newcommand{\QBars}[2]{%
  \fill[datagray!75] ({#1-0.28-0.19},{#2-0.46}) rectangle ({#1-0.28+0.19},{#2+0.88});
  \fill[\SpeechBarColor] ({#1+0.28-0.19},{#2-0.46}) rectangle ({#1+0.28+0.19},{#2+0.02});}

\newcommand{\ComparePanel}[2]{%
  \draw[panel] ({#1-\CompareHalf},{#2-\CompareHalf}) rectangle ({#1+\CompareHalf},{#2+\CompareHalf});
  \node[plab] at (#1,#2) {$D(\mathbf{p},\mathbf{q})$};}
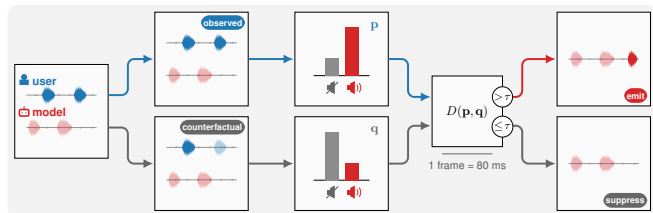
\begin{figure}[t]
  \centering
  \resizebox{\columnwidth}{!}{%
\begin{tikzpicture}[x=1cm,y=1cm]
\def\xA{1.8}
\def\xB{5.7}
\def\xC{9.6}
\def\xD{13.1}
\def\xE{16.9}
\def\yM{2.9}
\def\yT{4.35}
\def\yB{1.45}

\fill[panelgray,rounded corners=8pt] (0.30,-0.08) rectangle (18.4,5.85);

\SourcePanel{\xA}{\yM}
\ObservedPanel{\xB}{\yT}{observed}{\ObservedBadgeCorner}
\MutedPanel{\xB}{\yB}{counterfactual}{\MutedBadgeCorner}
\ProbabilityPanel{\xC}{\yT}{p}{predblue}{\PBars{\xC}{\yT}}
\ProbabilityPanel{\xC}{\yB}{q}{datagray}{\QBars{\xC}{\yB}}
\ComparePanel{\xD}{\yM}
\FrameLatencyNote{\xD}{\yM}
\EmitPanel{\xE}{\yT}{emit}{\EmitBadgeCorner}
\SuppressPanel{\xE}{\yB}{suppress}{\SuppressBadgeCorner}

\coordinate (Auser) at ({\xA+\HalfPanel},{\yM+\LaneTop});
\coordinate (Amodel) at ({\xA+\HalfPanel},{\yM+\LaneBot});
\coordinate (ObsIn) at ({\xB-\HalfPanel},{\yT});
\coordinate (MutIn) at ({\xB-\HalfPanel},{\yB});
\coordinate (Pout) at ({\xC+\HalfPanel},{\yT});
\coordinate (Qout) at ({\xC+\HalfPanel},{\yB});
\coordinate (Pin)  at ({\xD-\CompareHalf},{\yM+\EqYOffset});
\coordinate (Qin)  at ({\xD-\CompareHalf},{\yM-\EqYOffset});
\coordinate (NeqRoot) at ({\xD+\CompareHalf+0},{\yM+\EqYOffset});
\coordinate (EqRoot)  at ({\xD+\CompareHalf+0},{\yM-\EqYOffset});
\coordinate (NeqOut)  at ({\xD+\CompareHalf+0.24},{\yM+\EqYOffset});
\coordinate (EqOut)   at ({\xD+\CompareHalf+0.24},{\yM-\EqYOffset});
\coordinate (EmitIn) at ({\xE-\HalfPanel},{\yT});
\coordinate (SupIn)  at ({\xE-\HalfPanel},{\yB});

\draw[flowc,draw=predblue]
  (Auser) -- ({\xA+\HalfPanel+\ArrowStub},{\yM+\LaneTop})
          -- ({\xA+\HalfPanel+\ArrowStub},{\yT})
          -- (ObsIn);
\draw[flowc,draw=datagray]
  (Amodel) -- ({\xA+\HalfPanel+\ArrowStub},{\yM+\LaneBot})
           -- ({\xA+\HalfPanel+\ArrowStub},{\yB})
           -- (MutIn);

\draw[flowc,draw=predblue] ({\xB+\HalfPanel},{\yT}) -- ({\xC-\HalfPanel},{\yT});
\draw[flowc,draw=datagray] ({\xB+\HalfPanel},{\yB}) -- ({\xC-\HalfPanel},{\yB});

\draw[flowc,draw=predblue]
  (Pout) -- ({\xC+\HalfPanel+\ArrowStub},{\yT})
         -- ({\xC+\HalfPanel+\ArrowStub},{\yM+\EqYOffset})
         -- (Pin);
\draw[flowc,draw=datagray]
  (Qout) -- ({\xC+\HalfPanel+\ArrowStub},{\yB})
         -- ({\xC+\HalfPanel+\ArrowStub},{\yM-\EqYOffset})
         -- (Qin);

\draw[flowc,draw=emitred]
  (NeqOut) -- ({\xD+\CompareHalf+\ArrowStub},{\yM+\EqYOffset})
           -- ({\xD+\CompareHalf+\ArrowStub},{\yT})
           -- (EmitIn);
\draw[flowc,draw=datagray]
  (EqOut) -- ({\xD+\CompareHalf+\ArrowStub},{\yM-\EqYOffset})
          -- ({\xD+\CompareHalf+\ArrowStub},{\yB})
          -- (SupIn);

\node[
  draw=black,
  circle,
  fill=white,
  line width=0.6pt,
  inner sep=0.5pt,
  minimum size=16pt,
  font=\fontsize{9}{9.2}\selectfont\bfseries
] at (NeqRoot) {$>\!\tau$};

\node[
  draw=black,
  circle,
  fill=white,
  line width=0.6pt,
  inner sep=0.5pt,
  minimum size=16pt,
  font=\fontsize{9}{9.2}\selectfont\bfseries,
  yshift=\EqTextYShift
] at (EqRoot) {$\leq\!\tau$};

\end{tikzpicture}}
\vspace{-1.5em}
\caption{\textbf{To suppress spurious speech in full-duplex LLMs, we compare each proposed onset against a matched counterfactual.}
The observed continuation receives the newly arrived user audio, while the counterfactual continuation replaces the corresponding frames with silence; both preserve the preceding conversation and generated model history. {\mycolor{blue} We emit the proposed lexical onset when the resulting divergence exceeds a selected threshold and suppress it otherwise.} The comparison occurs within one 80-ms frame.}

\vspace{-0.5em}
  \label{fig:counterfactual}
\end{figure}

Full-duplex language modeling aims to address these challenges by absorbing speech recognition, language generation, and speech synthesis into a single learned model that processes incoming audio while generating its own output. For example, SyncLLM represents the user and model as synchronized language chunks~\cite{veluri2024syncllm}, while Moshi and PersonaPlex jointly advance user audio, model text, and model audio streams under a shared autoregressive clock~\cite{defossez2024moshi,roy2026personaplex}. As a result, these models can respond before the user finishes speaking, react to interruptions, coordinate overlapping speech, and preserve vocal information that would be discarded by transcription.

While elegant, delegating responsibility for speech onset to the generative model introduces a new failure mode. Specifically, we find that after either Moshi or PersonaPlex finishes answering a user and the user remains completely silent, the model keeps advancing its synchronized streams; after several seconds or minutes of nonlexical output, it often breaks the silence by hallucinating a new utterance despite receiving no new user speech. Controlling these onsets is difficult because they arise from the very continuous generative process that enables timely responses, interruptions, and overlapping conversation. Of course, turn-based gating could suppress model speech after a sufficiently long silence; however, the duration of preceding silence cannot determine whether the next onset arises autonomously or resumes a response to the user. For example, a full-duplex model may acknowledge a difficult question before pausing momentarily and continuing with its answer, thereby affecting latency measures such as ``retrieval delay'' and ``end-to-end keyword delay''~\cite{chien2026moshirag}. Imagine, though, that this pause outlasts the fixed activity window. In such a scenario, the gate would suppress the continued answer, even though the answer remains a response to the user's request. Gates such as these return responsibility for speech timing to an external turn detector, defeating the purpose of full-duplex modeling.

What causes {\mycolor{blue} these aforementioned full-duplex models} to escape silence when they should stay silent? Even when the user supplies no new speech, Moshi and PersonaPlex {\mycolor{brown} continue sampling model outputs, appending them to the generated history and conditioning each subsequent prediction on that history.} Continuing to build history during silence may produce spurious onsets in two ways: {\mycolor{brown} small probabilities of lexical onset may accumulate across repeated samples until the model begins speaking, or alternatively, self-conditioning on the generated nonlexical outputs may change the model's activations, causing an abrupt spike in onset probability.} These competing hypotheses imply different remedies: {\mycolor{brown} if onset comes from repeatedly sampling a small lexical tail, a decoding rule such as top-$k$ could remove that tail before it is ever sampled; meanwhile, if the model instead assigns nearly all probability to lexical onset at certain frames, restricting low-probability tokens cannot prevent it, and a more elaborate method is needed to distinguish desirable and undesirable onsets. We therefore investigate whether onset probability remains small throughout silence, or spikes immediately before speech.}

\textbf{This work.} In this paper, we first survey recent developments in speech-to-speech language modeling~(\Sref{sec:related}). {\mycolor{brown} Then, we empirically document spurious speech onset across two publicly available frontier full-duplex models, Moshi and PersonaPlex, and distinguish whether continued history construction produces these onsets through accumulated sampling from small probabilities or spikes caused by self-conditioning~(\Sref{sec:inspection}).}
Next, having explained how {\mycolor{blue} these models} escape silence, we develop a method for deciding which lexical onsets should be emitted or suppressed~(\Sref{sec:mitigation}).
Finally, we evaluate this intervention on {\mycolor{turquoise} held-out rollouts containing a genuine response followed by user silence} and find that it successfully suppresses {\mycolor{purple} spurious onsets while allowing desirable ones}~(\Sref{sec:results}). {\mycolor{brown} Not only do we arrive at a scientific understanding of the failure, but we also derive an actionable intervention from it.}

\section{Related Work}
\label{sec:related}

\textbf{\indent Streaming cascades.}
{\mycolor{blue} Before end-to-end full-duplex models, incremental spoken dialogue systems sought to reduce latency and relax rigid turn boundaries by closely chaining recognition, dialogue, and synthesis components. For example, NUMBERS propagated partial updates across every component, allowing the system to provide feedback while the user spoke and react to feedback while producing its own speech~\cite{skantze2009incremental}. More recently, NeuralFSM and DuplexCascade use language models to emit control tokens that determine whether a modular system should wait, respond, or interrupt~\cite{wang2024fullduplex,yang2026duplexcascade}. Systems in this category are commonly used in industry settings, including to automate customer service calls at DoorDash and GE Appliances and to process voice commands at Comcast~\cite{awsdoordash,awslexcustomers,tang2026cascading}. Though useful, these pipelines are inherently limited by the fact  that recognition, language generation, and synthesis are separate modules.}

\textbf{Full-duplex spoken language modeling.}
To place listening and speaking within one generative process, recent spoken language models jointly represent user and model streams. For instance, SyncLLM represents dialogue as synchronized user and model language chunks and predicts model chunks while new user chunks arrive~\cite{veluri2024syncllm}. Whereas SyncLLM operates over language chunks, Moshi jointly advances user-audio, model-text, and model-audio streams under a shared autoregressive clock~\cite{defossez2024moshi}. Furthermore, directly extending and fine-tuning Moshi, PersonaPlex simultaneously improves pause-handling behaviors and adds the ability to specify conversational personas~\cite{roy2026personaplex}. {\mycolor{blue} There also exist closed systems such as GPT-Live~\cite{openaigptlive} and Gemini Live~\cite{googlegeminilive}, but they do not expose weights for research purposes.} Alongside these developments, researchers have proposed benchmarks for interruption handling~\cite{lin2025fullduplexbench,ohashi2026interactivity}, overlapping speech~\cite{veluri2024syncllm,lin2025overlap}, backchanneling~\cite{lin2025fullduplexbench,ohashi2026interactivity}, and changes in speaker activity~\cite{lin2025fullduplexbench}.

\textbf{Free-running autoregressive generation.}
At inference time, an autoregressive model conditions each new prediction on its own previously sampled outputs. Accordingly, researchers studying exposure bias characterize the mismatch between teacher-forced training prefixes and free-running sampled prefixes~\cite{schmidt2019generalization,bengio2015scheduled,lamb2016professor,ross2011reduction}. The failure studied here is related in that self-conditioning can drive generated speech away from its ordinary response trajectory over long rollouts.

\textbf{{\mycolor{brown} Causal analysis through controlled interventions.}}
Those seeking {\mycolor{brown} causal explanations} of complex model behaviors distinguish {\mycolor{brown} between competing accounts} by holding most of a computation fixed while intervening on one candidate source of evidence~\cite{nanda2023progress,li2023emergent,nishi2025representation}. For instance, matched counterfactuals can test whether a model output depends on a particular feature or input. In full-duplex speech, the relevant source of evidence is user audio. {\mycolor{brown} We therefore apply this approach} by comparing the model's next-token distribution under observed user audio and under counterfactual silent audio.

\section{Inspecting Trained Moshi and PersonaPlex}
\label{sec:inspection}

\begin{figure}[t]
    \centering
    \includegraphics[width=\linewidth]{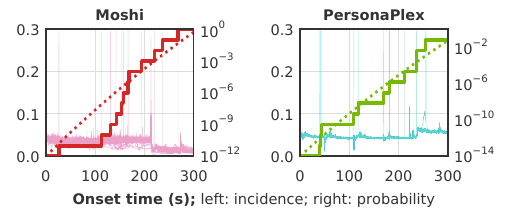}
\vspace{-1.5em}
    \caption{{\mycolor{brown} \textbf{Spurious onsets follow abrupt spikes in onset probability during self-conditioned generation.}}
    {\mycolor{brown} In each panel, the thick step curve shows cumulative onset incidence~(left axis), the dotted curve shows the corresponding constant-probability prediction, and the thin curves show frame-level onset probabilities from all 40 continuations~(logarithmic right axis). Onset probability remains near zero before spiking to nearly one at every observed onset; these spikes account for the steps in cumulative incidence.}}
\vspace{-0.5em}
    \label{fig:inspection}
\end{figure}

\textbf{\indent Experimental setup.}
Throughout, we use the official {\mycolor{blue} PyTorch BF16 7B checkpoints for Moshi and PersonaPlex}, the models' native 12.5-Hz frame rate, text sampling temperature $0.7$ with top-$k=25$, and audio sampling temperature $0.8$ with top-$k=250$. Text token IDs below 4 are padding or control outputs. {\mycolor{purple} As for how we classify a frame as a new onset, we first record 40 calibration responses per checkpoint to the fixed spoken request ``{\mycolor{blue} Tell me one funny joke right now in one sentence}'' and identify the final lexical frame of each response. The longest nonlexical gaps within the Moshi and PersonaPlex responses span 231 and 13 frames, respectively. Then, we define an onset to be a lexical token following one additional nonlexical frame.}
In these experiments, we reset all model and random-number-generator states; generate a deterministic greeting under seed 0; play the fixed spoken request; and allow the model to complete its response. {\mycolor{brown} To test whether a new user event is necessary to trigger spurious speech or whether it can arise completely autonomously as we suspect, we reseed the decoder and supply 3,750 digital-zero PCM frames, corresponding to five minutes. During this interval, the input audio is completely devoid of any audio events.} {\mycolor{turquoise} We measure time from when the model completes its first response, as determined by the calibrated nonlexical gap.}

\textbf{Onsets under silence.}
In \Fref{fig:inspection}, we plot the empirical cumulative incidence of lexical onsets across 40 five-minute continuations per model under uninterrupted digital-zero input. {\mycolor{purple} Moshi and PersonaPlex produce lexical onsets in {\mycolor{resultgreen} 30.0\%} and {\mycolor{resultgreen} 27.5\%} of continuations, respectively. The corresponding two-sided exact $95\%$ confidence intervals are {\mycolor{resultgreen} 16.6--46.5\%} and {\mycolor{resultgreen} 14.6--43.9\%}. PersonaPlex's median onset occurs after {\mycolor{resultgreen} 169.76} seconds rather than Moshi's {\mycolor{resultgreen} 161.96} seconds, {\mycolor{blue} showing that PersonaPlex exhibits less frequent and later spurious speech but still fails to avoid spurious speech completely.}} Considering that every frame of user audio contains digital zero in this controlled setting, no new acoustic or conversational event can be held responsible for these onsets. We therefore reject the account under which {\mycolor{blue} either model} begins a new utterance only in response to a clear triggering event.

\textbf{Distribution of onset times.} {\mycolor{purple} For these models, onset times span {\mycolor{resultgreen} 26.56} to {\mycolor{resultgreen} 268.00} seconds.} {\mycolor{brown} If either model consistently assigned a small probability of lexical onset at each frame, repeated sampling would begin accumulating incidence immediately, and the waiting time to first onset would follow a geometric distribution or an exponential distribution under a continuous-time approximation. However, the observed curves underrepresent onsets at earlier times before rising later, departing from the corresponding constant-hazard distributions in \Fref{fig:inspection}. We therefore hypothesize that onset probability varies both as individual histories develop and across histories.}

\textbf{Extracting history-dependent onset probabilities.}
{\mycolor{brown} Can we directly determine whether onset probability remains small throughout silence or spikes immediately before speech? Indeed, we can:} at every frame $t$ of continuation $i$, {\mycolor{blue} each model} produces the temperature-scaled, top-$k$ next-token distribution used for sampling. Letting $\mathcal{L}$ denote the set of lexical token IDs, the probability of lexical onset is
\begin{equation}
    h_{i,t}
    =
    \sum_{v\in\mathcal{L}}
    p_{i,t}(v),
    \label{eq:onset-hazard}
\end{equation}
where $p_{i,t}(v)$ is the probability assigned to token $v$ at frame $t$. {\mycolor{brown} For continuation $i$, the probability of sampling at least one lexical token before frame $T$, conditional on its evolving nonlexical history, is
\begin{equation}
    C_i(T)
    =
    1-\prod_{t=1}^{T-1}(1-h_{i,t}),
    \label{eq:cumulative-onset}
\end{equation}
before frame $T$. If small probabilities accumulate, then $C_i(T_{\mathrm{onset}})$ should become appreciable before onset. If self-conditioning instead produces a spike, then both $h_{i,t}$ and $C_i(t)$ should remain near zero until the onset frame. Consistent with the latter account, every observed onset follows an abrupt spike. Across all 23 onsets, $h_{i,t}$ is at most {\mycolor{resultgreen} $3.46\times10^{-10}$} on the immediately preceding frame and at least {\mycolor{resultgreen} $0.960$} on the onset frame. Correspondingly, $C_i(T_{\mathrm{onset}})$ never exceeds {\mycolor{resultgreen} $3.40\times10^{-7}$}. Thus, neither model escapes silence by eventually sampling a lexical token assigned a persistently small probability. Instead, self-conditioning causes onset probability to spike by more than nine orders of magnitude within one frame, after which lexical speech is nearly certain. Because the proposed onset already dominates the sampled distribution, removing low-probability tokens through top-$k$ cannot suppress the failure. A more elaborate intervention is needed.}

\section{Mitigating Spurious Speech}
\label{sec:mitigation}

So far, we have shown that {\mycolor{brown} self-conditioning during silence can produce abrupt spikes in onset probability.} Can we simply suppress all spikes? The answer is a definite no, since {\mycolor{brown} genuine responses should also produce spikes in onset probability when the model decides to speak at an appropriate time. Spikes indicate \textit{when} the model intends to speak, but they cannot tell us \textit{why}.} We must instead distinguish whether each proposed onset depends on the user's incoming request.

\textbf{{\mycolor{blue} Distinguishing between user-dependent and spurious onsets.}}
For each lexical onset, there is a simple question that one can ask, the answer to which we use to distinguish {\mycolor{blue} responses that depend on the user's inputted speech from spurious onsets}. That is this: \textit{Does the model's spoken response depend on the user's most recent speech input, or would the model predict a {\mycolor{blue} similar} next-token distribution even if that audio were muted?} {\mycolor{blue} If the response causally depends on this user input, we classify it as genuine; otherwise, as spurious.} We can implement this idea by forking histories from the frame immediately preceding that user utterance, saving the model's streaming state, and constructing two continuations, as illustrated in \Fref{fig:counterfactual}. The observed continuation receives the user audio, while the counterfactual continuation receives the corresponding frames without user speech---we can then compare between the model's predictions given each of the forked continuations.
{\mycolor{blue} As an aside, the position of the forking frame is known from the request waveform in our controlled experiments; online, a streaming voice activity detector (VAD) may identify the frames of user speech, and the decoder may branch from the preceding state. We emphasize that the VAD here may select the counterfactual forking point, but critically, unlike in cascaded systems with explicit turns, the full-duplex model still itself decides when it wants to speak.}

\textbf{Formalization.}
Let $p_t$ denote the model's next-token distribution for a lexical proposal at frame $t$ under the observed user audio, and let $q_t$ denote the corresponding distribution when that audio is muted. We condition both predictions on the generated model-text history through frame $t-1$, such that they differ only in the supplied user audio. {\mycolor{blue} Concretely, the observed continuation samples model-text tokens, the counterfactual continuation is teacher-forced with those tokens, and each continuation generates its own model-audio tokens under its assigned user audio.} To compare these distributions symmetrically, we define their midpoint $m_t=(p_t+q_t)/2$ and measure their Jensen--Shannon divergence,
\begin{equation}
    D_t
    =
    \frac{1}{2}\operatorname{KL}(p_t\|m_t)
    +
    \frac{1}{2}\operatorname{KL}(q_t\|m_t).
    \label{eq:counterfactual-score}
\end{equation}
{\mycolor{blue} Once a threshold $\tau$ is selected, we suppress lexical proposals with $D_t\leq\tau$ and preserve those with $D_t>\tau$. We study how this choice trades off suppression of spurious onsets against preservation of genuine responses in \Sref{sec:results}.}

\section{Experiments and Results}
\label{sec:results}

\textbf{\indent Threshold selection.}
{\mycolor{blue} {\mycolor{turquoise} Using 100 of 500 random rollouts per model, we sweep decision thresholds by comparing the first genuine response onset near the start of each rollout with the first spurious onset that occurs within a five-minute window of user silence following the end of the genuine response.} We select three thresholds: the smallest that preserves every genuine response without failing to suppress an additional spurious onset, the midpoint of the equivalence interval that maximizes the mean response-preservation and onset-suppression rate, and the largest that suppresses every spurious onset without suppressing an additional genuine response. We evaluate these ``conservative,'' ``balanced,'' and ``aggressive'' threshold points on the remaining 400 held-out rollouts.}

\begin{figure}[t]
    \centering
    \includegraphics[width=\linewidth]{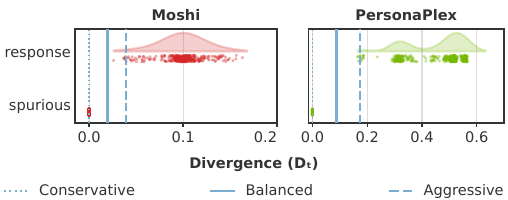}
    \vspace{-1.5em}
    \caption{\textbf{A threshold on distributional divergence cleanly separates spurious onsets from genuine responses.}
    {\mycolor{purple} Spurious onsets generally produce smaller distributional changes than genuine responses across both models, with well-separated held-out distributions. Vertical lines mark the conservative, balanced, and aggressive thresholds calibrated on 100 rollouts per model. Low-density tails are omitted for display.}}
    \label{fig:decision}
\end{figure}

\textbf{Evaluation conditions.}
{\mycolor{blue} To simulate realistic microphone conditions, we use the BRD2601 Background Noise recording, an audio clip of silence captured by the development board's microphone, as idle input and mix it into spoken requests~\cite{siliconlabsbrd2601}.} {\mycolor{purple} The noise and speech waveforms have RMS levels of $-96.0$ and $-19.1$ dBFS, respectively.

\textbf{Onset suppression and preservation results.}
In \Fref{fig:decision}, we show that the held-out scores separate remarkably cleanly: every spurious onset receives a smaller score than every genuine response. For spurious onsets, the model makes similar predictions across the two branches; for genuine responses, however, muting the user's speech changes those predictions. The balanced threshold suppresses spurious onsets with one-sided $95\%$ exact binomial lower confidence bounds of {\mycolor{resultgreen} 98.68\%} for Moshi and {\mycolor{resultgreen} 96.90\%} for PersonaPlex, while preserving genuine responses with bounds of {\mycolor{resultgreen} 98.82\%} for Moshi and {\mycolor{resultgreen} 99.25\%} for PersonaPlex~(\Tref{tab:results}). The threshold can be chosen according to which error is more costly: higher thresholds prioritize suppressing spurious onsets at the risk of rejecting genuine responses, while lower thresholds prioritize preserving genuine responses at the risk of allowing spurious onsets.}

\textbf{Generalizing beyond Moshi to PersonaPlex.}
{\mycolor{blue} We are particularly interested in PersonaPlex, as the authors claim it improves upon Moshi's pause handling via fine-tuning~\cite{roy2026personaplex}.} {\mycolor{purple} Our results verify that PersonaPlex is indeed better behaved, but only marginally: it initiates speech in {\mycolor{resultgreen} 27.5\%} rather than {\mycolor{resultgreen} 30.0\%} of continuations under digital-zero input, and its median onset occurs after {\mycolor{resultgreen} 169.76} rather than {\mycolor{resultgreen} 161.96} seconds. Despite the extra fine-tuning PersonaPlex underwent, it still escapes silence frequently and requires inference-time mitigation in practice. Earlier, we showed that {\mycolor{blue} our intervention substantially suppresses} spurious onsets while preserving desirable responses for both Moshi and PersonaPlex; that confirms that our mitigation method is not specific to certain Moshi checkpoints and transfers well.}

\begin{table}[t]
    \centering
    \setlength{\tabcolsep}{2pt}
    \normalsize
    \newcommand{\tablepercent}{\makebox[1em][r]{\%}}
    \begin{tabular}{l|c||c|c|c}
        \toprule
        & Spurious & & Spurious & Genuine \\[-1.5pt]
        Model & Onset & Threshold & Onsets & Responses \\[-1.5pt]
        & Count & & Suppressed & Preserved \\
        \midrule
        \multirow[c]{3}{*}{Moshi} & \multirow[c]{3}{*}{225} & Conservative & {\mycolor{resultgreen} 98.68\tablepercent} & {\mycolor{resultgreen} 98.82\tablepercent} \\
        & & \textbf{Balanced} & {\mycolor{resultgreen} \textbf{98.68\tablepercent}} & {\mycolor{resultgreen} \textbf{98.82\tablepercent}} \\
        & & Aggressive & {\mycolor{resultgreen} 98.68\tablepercent} & {\mycolor{resultgreen} 96.42\tablepercent} \\
        \midrule
        \multirow[c]{3}{*}{PersonaPlex} & \multirow[c]{3}{*}{95} & Conservative & {\mycolor{resultgreen} 96.90\tablepercent} & {\mycolor{resultgreen} 99.25\tablepercent} \\
        & & \textbf{Balanced} & {\mycolor{resultgreen} \textbf{96.90\tablepercent}} & {\mycolor{resultgreen} \textbf{99.25\tablepercent}} \\
        & & Aggressive & {\mycolor{resultgreen} 96.90\tablepercent} & {\mycolor{resultgreen} 97.73\tablepercent} \\
        \bottomrule
    \end{tabular}
    \vspace{-0.5em}
    \caption{\textbf{Our mitigation can suppress spurious onsets while preserving genuine ones.} It is effective for both Moshi and PersonaPlex. Percentages are one-sided $95\%$ lower confidence bounds.}
    \label{tab:results}
\vspace{-0.5em}
\end{table}

\textbf{Running in real-time.}
{\mycolor{purple} To run in real-time, our method must complete all computations within one frame. Indeed, every counterfactual comparison completes within the required 80-ms frame at the 95th percentile.} We carried out this benchmark by waiting for 100 warm-up frames, then measuring 1,000 {\mycolor{blue} paired forward steps} on one NVIDIA RTX PRO 6000 96GB GPU. {\mycolor{purple} The resulting 95th-percentile decision times are {\mycolor{resultgreen} 45.60} and {\mycolor{resultgreen} 60.61} ms for Moshi and PersonaPlex, respectively, below the 80-ms frame budget.}

\FloatBarrier

\section{Limitations}
\label{sec:limitations}

\textbf{\indent Conversational and acoustic conditions.}
We tested one spoken request and microphone noise from one device. Other questions, voices, microphones, and environments may produce different model predictions, so broader evaluations across conversational and recording conditions may be needed to tune the method for deployment.

\textbf{Selecting user input to mute.}
Suppose a user says ``Speak in ten seconds; ignore me until then.'' The model replies ``Got it,'' and the user keeps talking. Muting the most recent speech leaves the earlier request intact, so our method could suppress the delayed response. Accordingly, selecting the forking point may require reasoning across turns, beyond VAD. We focus on practical, training-free interventions on existing models; future work could explore a small trained selector.

\textbf{Generalizability across full-duplex architectures.}
Although PersonaPlex changes Moshi's voice and conversational role, it retains Moshi's organization of user audio, Inner Monologue, and model audio~\cite{defossez2024moshi,roy2026personaplex}. Other full-duplex systems use substantially different representations and generation procedures~\cite{veluri2024syncllm,yang2026duplexcascade}, and such differences may change both how spurious speech arises and how it can be suppressed. As new full-duplex architectures emerge, we recommend studying each architecture on its own terms and asking {\mycolor{brown} whether our explanation of spurious speech and our proposed intervention apply}. At least for the time being, our results establish the spurious speech phenomenon and our mitigation across both Moshi and PersonaPlex.

\textbf{Generalizability to asynchronous tool interactions.}
{\mycolor{brown} Our counterfactual analysis} and intervention both assume that new external input arrives only through the user's microphone. However, modern agentic systems also receive tool outputs and retrieved context~\cite{chien2026moshirag,qwenaudioagent}. For example, Qwen Audio Agent allows conversation to continue while a background agent executes tasks and returns results to the ongoing context~\cite{qwenaudioagent}. These rapidly evolving architectures are not always publicly available; Qwen Audio Agent relies on the hosted Qwen Audio 3.0 Realtime model, whose weights have not been released at the time of writing. We therefore study open speech models without committing to any particular tool interface. Future work should extend our counterfactual comparisons across tool and model states to distinguish onsets caused by user audio from those caused by newly returned information. For the underlying speech models, however, our results nevertheless distinguish spurious from desirable speech through causal dependence on external microphone input.

\section{Conclusion}
\label{sec:conclusion}

{\mycolor{brown} In summary, we show that under prolonged user silence, Moshi and PersonaPlex initiate spurious speech~(\Sref{sec:inspection}). By supplying digital-zero input, we remove new user-supplied events and show that these onsets arise autonomously rather than in response to an acoustic trigger~(\Sref{sec:inspection}). Then, we show that Moshi and PersonaPlex escape silence because self-conditioning on generated history causes the probability of lexical onset to spike, as evidenced by our finding that this probability increases by more than nine orders of magnitude at every observed onset and that these spikes, occurring at different times across continuations, produce the gradual cumulative incidence in \Fref{fig:inspection}. This result rejects accumulated sampling from small probabilities and shows why pruning a low-probability sampling tail cannot prevent the failure. However, genuine responses also produce spikes, so a spike identifies when the model intends to speak but cannot determine whether that speech depends on the user.} We therefore compare each proposed onset against the distribution obtained when the most recent user audio is muted~(\Sref{sec:mitigation}). {\mycolor{purple} Across held-out trials with Moshi and PersonaPlex, this intervention suppresses spurious onsets while preserving responses to user speech under realistic microphone noise} {\mycolor{brown} while running in real-time}{\mycolor{purple}~(\Sref{sec:results}).}

Taken together, our work advances both understanding and methodology pertaining to full-duplex models. We hope our contributions help make these models more reliable without sacrificing the free-flowing interactions they enable. As this modeling paradigm evolves, we invite researchers to trace failures in speech timing to the mechanisms that produce them and design interventions accordingly. We look forward to systems whose conversational flexibility is matched by principled control over when and why they speak.

\clearpage
\section{Compliance with Ethical Standards}

We evaluate full-duplex models under controlled audio inputs. We do not involve human participants, record private conversations, or use personally identifiable data; as such, no ethical approval was required.

Our goal is to make full-duplex voice agents speak more reliably, but that does not ensure the generated speech is factual, safe, or fair. Furthermore, calibration under one set of voices, microphones, and acoustic conditions may result in biased behaviors. We therefore recommend participatory engagement with real users and environments, maintaining independent safeguards for response content, and minimizing retention of user data. Finally, since improved speech timing may make automated agents appear more human, developers should clearly disclose that users are interacting with a synthetic agent.

\section{Funding Acknowledgement}

Comcast provided funding and computational resources during an internship under Akshat Pandey, Karun Kumar, and Tony Braskich.

\bibliographystyle{IEEEbib}
\bibliography{refs}

@article{yang2026duplexcascade,
  author = {Jianing Yang and Yusuke Fujita and Yui Sudo},
  title = {DuplexCascade: Full-Duplex Speech-to-Speech Dialogue with {VAD}-Free Cascaded {ASR}-{LLM}-{TTS} Pipeline and Micro-Turn Optimization},
  journal = {arXiv preprint arXiv:2603.09180},
  year = {2026},
  url = {https://arxiv.org/abs/2603.09180}
}

@inproceedings{skantze2009incremental,
  author = {Gabriel Skantze and David Schlangen},
  title = {Incremental Dialogue Processing in a Micro-Domain},
  booktitle = {Proceedings of the 12th Conference of the European Chapter of the Association for Computational Linguistics},
  pages = {745--753},
  year = {2009},
  publisher = {Association for Computational Linguistics},
  doi = {10.3115/1609067.1609150}
}

@inproceedings{wang2024fullduplex,
  author = {Peng Wang and Songshuo Lu and Yaohua Tang and Sijie Yan and Wei Xia and Yuanjun Xiong},
  title = {A Full-Duplex Speech Dialogue Scheme Based on Large Language Model},
  booktitle = {Advances in Neural Information Processing Systems},
  volume = {37},
  pages = {13372--13403},
  year = {2024},
  doi = {10.52202/079017-0427}
}

@inproceedings{veluri2024syncllm,
  author = {Bandhav Veluri and Benjamin N. Peloquin and Bokai Yu and Hongyu Gong and Shyamnath Gollakota},
  title = {Beyond Turn-Based Interfaces: Synchronous {LLM}s as Full-Duplex Dialogue Agents},
  booktitle = {Proceedings of the 2024 Conference on Empirical Methods in Natural Language Processing},
  pages = {21390--21402},
  year = {2024},
  publisher = {Association for Computational Linguistics},
  doi = {10.18653/v1/2024.emnlp-main.1192}
}

@article{defossez2024moshi,
  author = {Alexandre D{\'e}fossez and Laurent Mazar{\'e} and Manu Orsini and Am{\'e}lie Royer and Patrick P{\'e}rez and Herv{\'e} J{\'e}gou and Edouard Grave and Neil Zeghidour},
  title = {Moshi: A Speech-Text Foundation Model for Real-Time Dialogue},
  journal = {arXiv preprint arXiv:2410.00037},
  year = {2024},
  url = {https://arxiv.org/abs/2410.00037}
}

@inproceedings{roy2026personaplex,
  author = {Rajarshi Roy and Jonathan Raiman and Sang-Gil Lee and Teodor-Dumitru Ene and Robert Kirby and Sungwon Kim and Jaehyeon Kim and Bryan Catanzaro},
  title = {PersonaPlex: Voice and Role Control for Full Duplex Conversational Speech Models},
  booktitle = {2026 IEEE International Conference on Acoustics, Speech and Signal Processing (ICASSP)},
  pages = {16137--16141},
  year = {2026},
  publisher = {IEEE},
  doi = {10.1109/ICASSP55912.2026.11463413}
}

@inproceedings{lin2025fullduplexbench,
  author = {Guan-Ting Lin and Jiachen Lian and Tingle Li and Qirui Wang and Gopala Anumanchipalli and Alexander H. Liu and Hung-Yi Lee},
  title = {Full-Duplex-Bench: A Benchmark to Evaluate Full-Duplex Spoken Dialogue Models on Turn-Taking Capabilities},
  booktitle = {2025 IEEE Automatic Speech Recognition and Understanding Workshop (ASRU)},
  pages = {1--8},
  year = {2025},
  publisher = {IEEE},
  doi = {10.1109/ASRU65441.2025.11433838}
}

@inproceedings{lin2025overlap,
  author = {Guan-Ting Lin and Shih-Yun Shan Kuan and Qirui Wang and Jiachen Lian and Tingle Li and Shinji Watanabe and Hung-Yi Lee},
  title = {{Full-Duplex-Bench} v1.5: Evaluating Overlap Handling for Full-Duplex Speech Models},
  booktitle = {2026 IEEE International Conference on Acoustics, Speech and Signal Processing (ICASSP)},
  pages = {19447--19451},
  year = {2026},
  publisher = {IEEE},
  doi = {10.1109/ICASSP55912.2026.11463576}
}

@article{ohashi2026interactivity,
  author = {Atsumoto Ohashi and Neil Zeghidour and Alexandre D{\'e}fossez and Eugene Kharitonov},
  title = {Multi-Faceted Interactivity Alignment in Full-Duplex Speech Models},
  journal = {arXiv preprint arXiv:2606.11167},
  year = {2026},
  url = {https://arxiv.org/abs/2606.11167}
}

@article{chien2026moshirag,
  author = {Chung-Ming Chien and Manu Orsini and Eugene Kharitonov and Neil Zeghidour and Karen Livescu and Alexandre D{\'e}fossez},
  title = {{MoshiRAG}: Asynchronous Knowledge Retrieval for Full-Duplex Speech Language Models},
  journal = {arXiv preprint arXiv:2604.12928},
  year = {2026},
  url = {https://arxiv.org/abs/2604.12928}
}

@inproceedings{schmidt2019generalization,
  author = {Florian Schmidt},
  title = {Generalization in Generation: A Closer Look at Exposure Bias},
  booktitle = {Proceedings of the 3rd Workshop on Neural Generation and Translation},
  pages = {157--167},
  year = {2019},
  publisher = {Association for Computational Linguistics},
  doi = {10.18653/v1/D19-5616}
}

@inproceedings{bengio2015scheduled,
  author = {Samy Bengio and Oriol Vinyals and Navdeep Jaitly and Noam Shazeer},
  title = {Scheduled Sampling for Sequence Prediction with Recurrent Neural Networks},
  booktitle = {Advances in Neural Information Processing Systems},
  volume = {28},
  year = {2015},
  url = {https://proceedings.neurips.cc/paper/2015/hash/e995f98d56967d946471af29d7bf99f1-Abstract.html}
}

@inproceedings{lamb2016professor,
  author = {Alex M. Lamb and Anirudh Goyal and Ying Zhang and Saizheng Zhang and Aaron C. Courville and Yoshua Bengio},
  title = {Professor Forcing: A New Algorithm for Training Recurrent Networks},
  booktitle = {Advances in Neural Information Processing Systems},
  volume = {29},
  year = {2016},
  url = {https://proceedings.neurips.cc/paper/2016/hash/16026d60ff9b54410b3435b403afd226-Abstract.html}
}

@inproceedings{ross2011reduction,
  author = {St{\'e}phane Ross and Geoffrey J. Gordon and J. Andrew Bagnell},
  title = {A Reduction of Imitation Learning and Structured Prediction to No-Regret Online Learning},
  booktitle = {Proceedings of the Fourteenth International Conference on Artificial Intelligence and Statistics},
  series = {Proceedings of Machine Learning Research},
  volume = {15},
  pages = {627--635},
  year = {2011},
  publisher = {PMLR},
  url = {https://proceedings.mlr.press/v15/ross11a.html}
}

@inproceedings{nanda2023progress,
  author = {Neel Nanda and Lawrence Chan and Tom Lieberum and Jess Smith and Jacob Steinhardt},
  title = {Progress Measures for Grokking via Mechanistic Interpretability},
  booktitle = {The Eleventh International Conference on Learning Representations},
  year = {2023},
  publisher = {OpenReview.net},
  url = {https://openreview.net/forum?id=9XFSbDPmdW}
}

@inproceedings{li2023emergent,
  author = {Kenneth Li and Aspen K. Hopkins and David Bau and Fernanda Vi{\'e}gas and Hanspeter Pfister and Martin Wattenberg},
  title = {Emergent World Representations: Exploring a Sequence Model Trained on a Synthetic Task},
  booktitle = {The Eleventh International Conference on Learning Representations},
  year = {2023},
  publisher = {OpenReview.net},
  url = {https://openreview.net/forum?id=DeG07_TcZvT}
}

@inproceedings{nishi2025representation,
  author = {Kento Nishi and Rahul Ramesh and Maya Okawa and Mikail Khona and Hidenori Tanaka and Ekdeep Singh Lubana},
  title = {Representation Shattering in Transformers: A Synthetic Study with Knowledge Editing},
  booktitle = {Proceedings of the 42nd International Conference on Machine Learning},
  series = {Proceedings of Machine Learning Research},
  volume = {267},
  pages = {46525--46553},
  year = {2025},
  publisher = {PMLR},
  url = {https://proceedings.mlr.press/v267/nishi25a.html}
}

@misc{siliconlabsbrd2601,
  author = {{Silicon Laboratories}},
  title = {{BRD2601 Background Noise}},
  howpublished = {{MLTK} 0.20.0 documentation},
  note = {Accessed August 13, 2026},
  url = {https://siliconlabs.github.io/mltk/docs/python_api/datasets/audio/background_noise/brd2601.html}
}

@misc{qwenaudioagent,
  author = {{Qwen Team}},
  title = {{Qwen Audio Agent}},
  howpublished = {GitHub repository},
  note = {Accessed August 13, 2026},
  url = {https://github.com/QwenAudio/qwen-audio-agent}
}

@misc{awsdoordash,
  author = {{Amazon Web Services}},
  title = {Building a Generative {AI} Contact Center Solution for {DoorDash} Using {Amazon Bedrock}, {Amazon Connect Customer}, and {Anthropic's Claude}},
  howpublished = {Customer case study},
  note = {Accessed August 13, 2026},
  url = {https://aws.amazon.com/solutions/case-studies/doordash-bedrock-case-study/}
}

@misc{awslexcustomers,
  author = {{Amazon Web Services}},
  title = {{Amazon Lex} Customers},
  howpublished = {Customer case studies},
  note = {Accessed August 13, 2026},
  url = {https://aws.amazon.com/lex/customers/}
}

@misc{tang2026cascading,
  author = {Raphael Tang and Yajie Mao and Karun Kumar and Ferhan Ture},
  title = {Systems and Methods for Managing Cascading Models},
  year = {2026},
  howpublished = {U.S. Patent Application Publication US 2026/0120708 A1},
  url = {https://patents.google.com/patent/US20260120708A1/en}
}

@misc{openaigptlive,
  author = {{OpenAI}},
  title = {{GPT-Live} System Card},
  year = {2026},
  url = {https://deploymentsafety.openai.com/gpt-live/gpt-live.pdf}
}

@misc{googlegeminilive,
  author = {{Google DeepMind}},
  title = {{Gemini Audio}: Live Dialogue},
  note = {Accessed August 13, 2026},
  url = {https://deepmind.google/models/gemini-audio/live-dialogue/}
}

\end{document}